\documentclass[12pt]{article}

\usepackage{arxiv}

\usepackage[utf8]{inputenc} 
\usepackage[T1]{fontenc}    
\usepackage{hyperref}       
\usepackage{url}            
\usepackage{booktabs}       
\usepackage{amsfonts}       
\usepackage{nicefrac}       
\usepackage{microtype}      
\usepackage{lipsum}		
\usepackage{graphicx}
\usepackage{doi}

\usepackage{graphicx}
\usepackage{amssymb}
\usepackage{textcomp}
\usepackage{endnotes}
\usepackage{color}
\usepackage{lineno}
\usepackage{colortbl}
\usepackage[natbib,backend=bibtex]{biblatex}
\usepackage{caption}
\usepackage{subcaption}
\usepackage[inline]{enumitem}

\hypersetup{
 colorlinks,
 citecolor=black,
 linkcolor=black,
 urlcolor=blue}
  
\title{Using Game Engines and Machine Learning \\[0.3ex] to Create Synthetic Satellite Imagery \\[0.3ex] for a Tabletop Verification Exercise}

\author{Johannes Hoster \\ 
Berliner Hochschule f\"ur Technik \\
	\And
Sara Al-Sayed \\
Program on Science and Global Security, Princeton University \\
\And
Felix Biessmann \\
Berliner Hochschule f\"ur Technik \\
Einstein Center Digital Future, Berlin \\
\And
Alexander Glaser \\
Program on Science and Global Security, Princeton University \\
Einstein Center Digital Future, Berlin \\
\And
Kristian Hildebrand \\
Berliner Hochschule f\"ur Technik \\
\And
Igor Moric \\
Program on Science and Global Security, Princeton University \\
\And
Tuong Vy Nguyen \\
Berliner Hochschule f\"ur Technik \\
}

\renewcommand{\shorttitle}{Synthetic Satellite Imagery for a Tabletop Verification Exercise}

\begin{document}
\maketitle

\begin{abstract}
Satellite imagery is regarded as a great opportunity for citizen-based monitoring of activities of interest. Relevant imagery may however not be available at sufficiently high resolution, quality, or cadence---let alone be uniformly accessible to open-source analysts. This limits an assessment of the true long-term potential of citizen-based monitoring of nuclear activities using publicly available satellite imagery. In this article, we demonstrate how modern game engines combined with advanced machine-learning techniques can be used to generate synthetic imagery of sites of interest with the ability to choose relevant parameters upon request; these include time of day, cloud cover, season, or level of activity onsite. At the same time, resolution and off-nadir angle can be adjusted to simulate different characteristics of the satellite. While there are several possible use-cases for synthetic imagery, here we focus on its usefulness to support tabletop exercises in which simple monitoring scenarios can be examined to better understand verification capabilities enabled by new satellite constellations and very short revisit times.	
\end{abstract}


~








\section{Introduction}



Satellites have been used since the 1950s for Earth observation, first by governments with reconnaissance missions shrouded in secrecy, but increasingly now also by commercial providers and by a growing community of open-source analysts. There are currently about 700 imaging satellites in orbit, and some constellations realize revisit times on the order of 20 minutes. Maxar Technologies, one of the main commercial satellite operators, currently acquires an average of about 80 terabytes of imagery per day. Over the next few years, these and other collection efforts will grow to an immense archive of digital data, and there is a widely shared expectation---or hope---that broad and open access to open-source information will enable the early or timely detection of non-compliance with relevant international agreements.
Here, we are particularly interested in monitoring compliance with nuclear nonproliferation and arms-control agreements, but similar opportunities are also emerging in the context of environmental and carbon-emission monitoring, emergency response and human-rights monitoring, and archaeological-site monitoring.
This article is part of an ongoing project that seeks to systematically assess the long-term potential of satellite imagery for monitoring and verification purposes.

\begin{figure}     
     \begin{subfigure}[b]{0.24\textwidth}
         \centering
         \includegraphics[width=\textwidth]{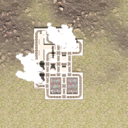}
         \caption{Game engine render\newline}
         \label{fig:MLECA3}
     \end{subfigure}
     \hfill
          \begin{subfigure}[b]{0.24\textwidth}
         \centering
         \includegraphics[width=\textwidth]{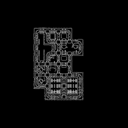}
         \caption{Input modality for structural guidance}
         \label{fig:MLESDL_overview}
     \end{subfigure}
     \hfill
          \begin{subfigure}[b]{0.24\textwidth}
         \centering
         \includegraphics[width=\textwidth]{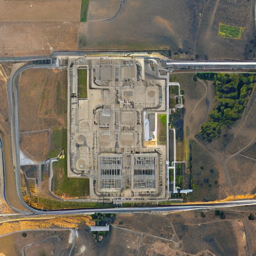}
         \caption{Synthesized image without details}
         \label{fig:MLESDL}
     \end{subfigure}
     \hfill
          \begin{subfigure}[b]{0.24\textwidth}
         \centering
         \includegraphics[width=\textwidth]{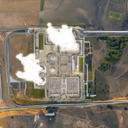}
         \caption{Synthesized image with reinserted details}
         \label{fig:Final}
     \end{subfigure}
        \caption{Overview of our method from game engine render, via control input modality for structural guidance, to reinsertion of details into the synthesized image}
        \label{fig:Overview}
\end{figure}

Working with real satellite imagery, however, has strong limitations. For one, access to high-resolution imagery can be extremely expensive when required at scale, i.e., on a daily basis for multiple sites or larger geographical regions. More importantly perhaps, the number of relevant existing facilities of interest is relatively small, making it more difficult to draw broader conclusions based on rather limited datasets. To overcome both these constraints, this project seeks to generate and leverage synthetic satellite imagery for notional sites showing various industrial sites or other human activities. This enables us to simulate the capabilities of arbitrary satellite constellations, while controlling relevant parameters of the imagery such as seasons, revisit times, or resolution and allowing for other gradual and abrupt changes. An overview of our method is illustrated in \autoref{fig:Overview}. We can then use this imagery to develop and examine concrete scenarios, in which parties are either compliant or non-compliant with an agreement.

\section{Related Work}

\paragraph{\textbf{Synthetic datasets supporting nuclear agreement monitoring:}} The first synthetic imagery datasets were mostly of the non-overhead kind and efforts for their generation were driven by autonomous driving R\&D, representative datasets being KITTI \cite{KITTI} and SYNTHIA \cite{SYNTHIA}. While there are many dedicated synthetic overhead (aerial or satellite) dataset generation efforts for different object classes (e.g., buildings, infrastructure such as roads, bridges, and dams, vehicles) or contexts of interest (e.g., urban scenes, agriculture, wildlife) and different computer vision applications (e.g., segmentation, classification, object detection and tracking, autonomous navigation, scene understanding, surveying, change detection, time-series analysis) -- representative datasets being Synthinel-1 \cite{Synthinel-1} and RarePlanes \cite{RarePlanes} (satellite) and UnityShip \cite{UnityShip} and BIRDSAI \cite{BIRDSAI} (aerial), to the best of our knowledge, there are none in the nuclear context, except the efforts by Gastelum et al. \cite{SafeguardsDataset} -- if not of the overhead kind -- and Nguyen et al. \cite{Nguyen2023}. The former represents work at the Sandia National Laboratories using a game engine to generate synthetic imagery of uranium hexafluoride containers -- part of the nuclear fuel cycle -- against various backgrounds typical of nuclear scenarios of interest, for computer vision applications. The work by Nguyen et al., on the other hand, focuses on text-conditioned synthetic satellite imagery generation of nuclear power plants. Apart from the latter work, there are no datasets serving the purpose of a comprehensive assessment of agreement monitoring using the current capabilities of commercial satellite constellations. That being said, synthetic data in the form of text, imagery, or video are leveraged or foreseen to be leveraged in the nuclear safeguards and nonproliferation context for specific applications, as suggested by the meeting agenda of the {\em Joint Technical Exchange on Synthetic Data for Safeguards and Nonproliferation: ESARDA Verification Technologies \& Methodologies and INMM Open Source and Geospatial Information Working Groups}, 2022, which was attended by the second author of this paper.

\paragraph{\textbf{Text-conditioned image generation:}} 
Text-to-image generation provides an intuitive and comprehensible way of conditional image synthesis, with large models like DALL-E 2 \cite{ramesh2022hierarchical} already demonstrating how powerful they can be. However, the issue with text-conditioned image generation alone is that there’s little control over the exact outcome. 
While there has been substantial progress in terms of image quality with text-conditioning for some aspects of images, controlling the output of image generation models precisely remains challenging. In this paper, we propose to first create a render of the model of interest, in this case a nuclear power plant, using the game engine then use feature maps like canny edge and depth map of the render as an additional guidance mechanism, which provides a way to further control the image generation process. 


\paragraph{\textbf{Applications for dataset generation:}}
Since computer vision models require a large amount of labeled data to be trained, generating synthetic data using machine learning techniques or game engines or a combination thereof is a simple and straightforward approach to create datasets. Unity Perception \cite{borkman2021unity}, for example, provides a highly customizable toolset for this task. Other applications are BlenderProc \cite{denninger2019blenderproc}, NVIDIA Isaac Gym \cite{makoviychuk2021isaac}, and NVISII \cite{morrical2021nvisii}.
All of these tools provide semantic or instance segmentation, bounding-box generation, and also, in some cases, support for other vision tasks such as depth maps, keypoints, normals, or optical flow.
However, our work doesn't focus on traditional machine vision tasks, such as distinguishing individual objects from their cluttered background, nor on understanding indoor scenes, but on a very specific task for which a dedicated kernel modelling algorithm is better suited.

\paragraph{\textbf{Game engines for system design, testing, and validation:}} While game engines in addition to being used to generate synthetic imagery are also being used to build simulators of aerial systems for surveillance \cite{OVVV} and autonomous navigation \cite{AirSim}, our tabletop exercise may inform the design, using game engines or other CAD tools, of the ideal satellite constellation for high-confidence global and comprehensive monitoring of nuclear activities that patches present technical gaps. Having imagery from the ideal satellite constellation at hand, one can properly assess the shortcomings of nuclear activity monitoring based on the current capabilities of contemporary commercial satellite constellations.

\section{Method}

We combine the strengths of game engines and machine-learning models to generate synthetic satellite images with configurable content and realistic appearance. Specifically, we use the Unity3D \cite{Unity} game engine to create a procedurally generated model of a nuclear power plant with randomizable parameters like camera distance and off-nadir angle, plant structure, time-of-day, cloud cover, and level of activity onsite. This provides the possibility to design the content of the created image as desired. 

\subsection{Imagery Generation Using Game Engine}
\begin{figure}
     \begin{subfigure}[b]{0.32\textwidth}
         \centering
         \includegraphics[width=\textwidth]{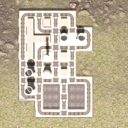}
         \caption{Basic structure}
         \label{fig:Act1}
     \end{subfigure}
        \hfill
     \begin{subfigure}[b]{0.32\textwidth}
         \centering
         \includegraphics[width=\textwidth]{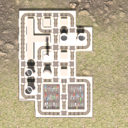}
         \caption{Cars added}
         \label{fig:Act2}
     \end{subfigure}
          \hfill
     \begin{subfigure}[b]{0.32\textwidth}
         \centering
         \includegraphics[width=\textwidth]{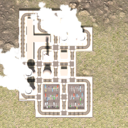}
         \caption{Steam added}
         \label{fig:Act3}
     \end{subfigure}
        \caption{Visualization of different components of the model from basic structure to high level of activity onsite}
        \label{fig:Activity}
\end{figure}

Since procedural level generation is widely used in the game industry, using a game engine such as Unity allows to create algorithms that can generate different layouts of pre-defined structures that typically make up a nuclear power plant, with customizable constraints. Since we employ the CoAdapter model later to make enhancements to the game engine render of the nuclear power plant model towards more realistic imagery, as explained in Section \ref{subsec:MLModel}, we focused more on the plant structure instead of the graphical appearance of the game engine render. That is, we created untextured models of the reactor, cooling towers, stacks, and tetromino-shaped buildings, and we chose a basic grid system to place the models in different layouts. Immediately after the plant model is constructed (typically in less than a second), details reflecting the level of activity onsite, such as the number of cars in the parking lot (\autoref{fig:Act2}) or the amount of steam emanating from the cooling towers (\autoref{fig:Act3}), can be added automatically. The density of these details can be preset or randomized to control the level of activity onsite.

\begin{figure}
     \hfill
          \begin{subfigure}[b]{0.19\textwidth}
         \centering
         \includegraphics[width=\textwidth]{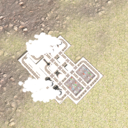}
         \caption{Off-nadir 10°}
         \label{fig:ON10}
     \end{subfigure}
     \hfill
     \begin{subfigure}[b]{0.19\textwidth}
         \centering
         \includegraphics[width=\textwidth]{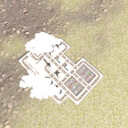}
         \caption{Off-nadir 20°}
         \label{fig:ON20}
     \end{subfigure}
     \hfill
          \begin{subfigure}[b]{0.19\textwidth}
         \centering
         \includegraphics[width=\textwidth]{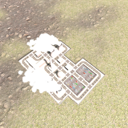}
         \caption{Off-nadir 30°}
         \label{fig:ON30}
     \end{subfigure}
     \hfill
     \begin{subfigure}[b]{0.19\textwidth}
         \centering
         \includegraphics[width=\textwidth]{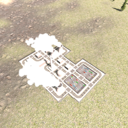}
         \caption{Off-nadir 40°}
         \label{fig:ON40}
     \end{subfigure}
     \hfill
     \begin{subfigure}[b]{0.19\textwidth}
         \centering
         \includegraphics[width=\textwidth]{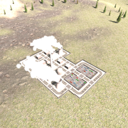}
         \caption{Off-nadir 50°}
         \label{fig:ON50}
     \end{subfigure}
        \caption{Visualization of different off-nadir angles}
        \label{fig:OffNadir}
\end{figure}

Since the encoded configurable model of the plant is 3-dimensional, the camera can be moved freely around the model. Thus, images could be created of several instantiations of the model with arbitrary off-nadir angles (\autoref{fig:OffNadir}).

\begin{figure}
     \begin{subfigure}[b]{0.24\textwidth}
         \centering
         \includegraphics[width=\textwidth]{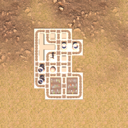}
         \caption{Morning}
         \label{fig:TimeM}
     \end{subfigure}
     \hfill
          \begin{subfigure}[b]{0.24\textwidth}
         \centering
         \includegraphics[width=\textwidth]{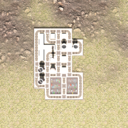}
         \caption{Day}
         \label{fig:TimeD}
     \end{subfigure}
     \hfill
          \begin{subfigure}[b]{0.24\textwidth}
         \centering
         \includegraphics[width=\textwidth]{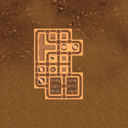}
         \caption{Evening}
         \label{fig:TimeE}
     \end{subfigure}
     \hfill
          \begin{subfigure}[b]{0.24\textwidth}
         \centering
         \includegraphics[width=\textwidth]{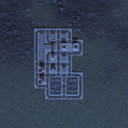}
         \caption{Night}
         \label{fig:TimeN}
     \end{subfigure}
        \caption{Visualization of different times-of-day}
        \label{fig:Time}
\end{figure}

\begin{figure}
     
     \begin{subfigure}[b]{0.24\textwidth}
         \centering
         \includegraphics[width=\textwidth]{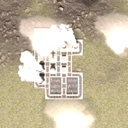}
         \caption{Low cloud coverage}
         \label{fig:CCL}
     \end{subfigure}
     \hfill
          \begin{subfigure}[b]{0.24\textwidth}
         \centering
         \includegraphics[width=\textwidth]{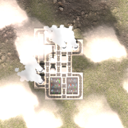}
         \caption{Medium cloud coverage}
         \label{fig:CCM}
     \end{subfigure}
     \hfill
          \begin{subfigure}[b]{0.24\textwidth}
         \centering
         \includegraphics[width=\textwidth]{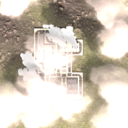}
         \caption{High cloud coverage}
         \label{fig:CCH}
     \end{subfigure}
     \hfill
          \begin{subfigure}[b]{0.24\textwidth}
         \centering
         \includegraphics[width=\textwidth]{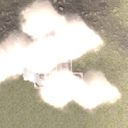}
         \caption{Extreme cloud coverage}
         \label{fig:CCE}
     \end{subfigure}
        \caption{Visualization of different cloud coverage}
        \label{fig:Cloud}
\end{figure}

Game engines also enable the rendering of models under different conditions such as the time-of-day (\autoref{fig:Time}) or with different cloud parameters (\autoref{fig:Cloud}).
Overall, our approach enables the creation of large high-variability datasets in a short time.

\subsection{Multimodally-guided Satellite Imagery Synthesis}
\label{subsec:MLModel}
Utilizing the game engine with virtual models of the relevant elements enables the generation of many images in a short time with different features. However, despite the extensive graphic achievements in recent years, the output of the engine alone still lacks realism to a large degree, impeding the successful execution of downstream tasks. 

\begin{figure}
     \begin{subfigure}[b]{0.32\textwidth}
         \centering
         \includegraphics[width=\textwidth]{figures/MLExample/Activity1CannyWB.png}
     \end{subfigure}
     \hfill
     \begin{subfigure}[b]{0.32\textwidth}
         \centering
         \includegraphics[width=\textwidth]{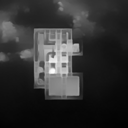}
     \end{subfigure}
     \hfill
     \begin{subfigure}[b]{0.32\textwidth}
         \centering
         \includegraphics[width=\textwidth]{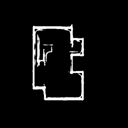}
     \end{subfigure}
     
     \begin{subfigure}[b]{0.32\textwidth}
         \centering
         \includegraphics[width=\textwidth]{figures/MLExample/Activity1CannyWBSatellite_image_of_a_nuclear_power_plant.png}
         \caption{Canny edge}
         \label{fig:MLEC}
     \end{subfigure}
     \hfill
          \begin{subfigure}[b]{0.32\textwidth}
         \centering
         \includegraphics[width=\textwidth]{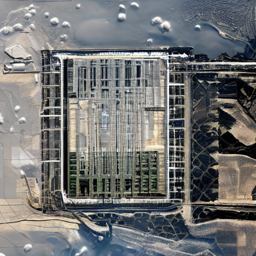}
         \caption{Depth map}
         \label{fig:MLED}
     \end{subfigure}
     \hfill
          \begin{subfigure}[b]{0.32\textwidth}
         \centering
         \includegraphics[width=\textwidth]{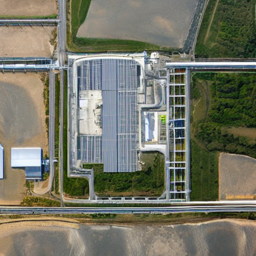}
         \caption{Sketch}
         \label{fig:MLES}
     \end{subfigure}
        \caption{Visualization of different input modalities (top) and corresponding results (bottom). In addition to the visual input modality, the text prompt ``Satellite image of a nuclear power plant'' was used.}
        \label{fig:MLExample}
\end{figure}

\begin{figure}
     \begin{subfigure}[b]{0.24\textwidth}
         \centering
         \includegraphics[width=\textwidth]{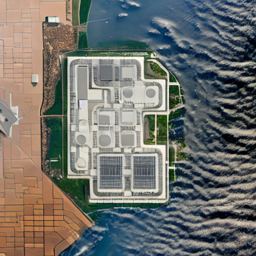}
         \caption{Canny + Depth}
         \label{fig:CD}
     \end{subfigure}
     \hfill
          \begin{subfigure}[b]{0.24\textwidth}
         \centering
         \includegraphics[width=\textwidth]{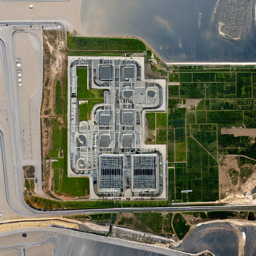}
         \caption{Canny + Sketch}
         \label{fig:CS}
     \end{subfigure}
     \hfill
          \begin{subfigure}[b]{0.24\textwidth}
         \centering
         \includegraphics[width=\textwidth]{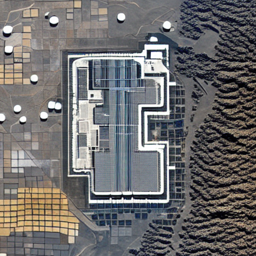}
         \caption{Depth + Sketch}
         \label{fig:DS}
     \end{subfigure}
     \hfill
          \begin{subfigure}[b]{0.24\textwidth}
         \centering
         \includegraphics[width=\textwidth]{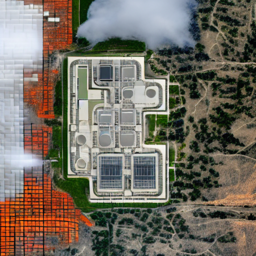}
         \caption{Canny + Depth + Sketch}
         \label{fig:CDS}
     \end{subfigure}
        \caption{Results from multiple different input modalities with text prompt, ``Satellite image of a nuclear power plant''}
        \label{fig:MultiInput}
\end{figure}

We solve this problem by feeding the images of the game engine into a T2I \cite{mou2023t2iadapter} Composable Adapter (CoAdapter) model \cite{T2I}, which is based on machine learning. The CoAdapter allows for multiple input modalities to define the appearance as well as content of the image.
The most promising input modalities to capture the layout of the input image are the canny edge, depth map, and sketch modalities, which can be seen along with the corresponding results in \autoref{fig:MLExample}. 
It is also possible to use multiple input modalities at once (\autoref{fig:MultiInput}).

\begin{figure}

          \begin{subfigure}[b]{0.24\textwidth}
         \centering
         \includegraphics[width=\textwidth]{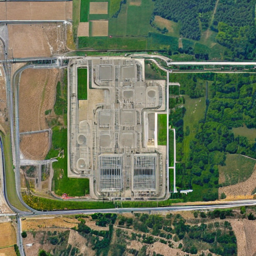}
     \end{subfigure}
     \hfill
     \begin{subfigure}[b]{0.24\textwidth}
         \centering
         \includegraphics[width=\textwidth]{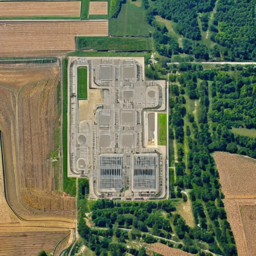}
     \end{subfigure}
     \hfill
     \begin{subfigure}[b]{0.24\textwidth}
         \centering
         \includegraphics[width=\textwidth]{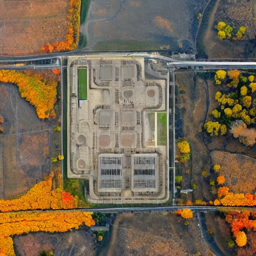}
     \end{subfigure}
     \hfill
     \begin{subfigure}[b]{0.24\textwidth}
         \centering
         \includegraphics[width=\textwidth]{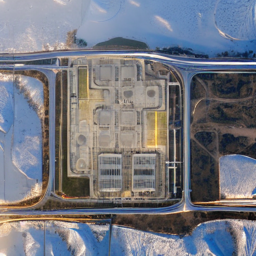}
     \end{subfigure}
     
     \begin{subfigure}[b]{0.24\textwidth}
         \centering
         \includegraphics[width=\textwidth]{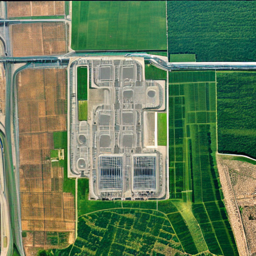}
         \caption{Spring}
         \label{fig:SSpGSH}
     \end{subfigure}
     \hfill
          \begin{subfigure}[b]{0.24\textwidth}
         \centering
         \includegraphics[width=\textwidth]{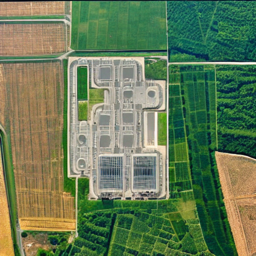}
         \caption{Summer}
         \label{fig:SSuGSH}
     \end{subfigure}
     \hfill
          \begin{subfigure}[b]{0.24\textwidth}
         \centering
         \includegraphics[width=\textwidth]{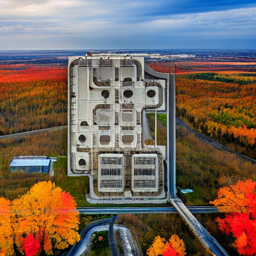}
         \caption{Fall}
         \label{fig:SFGSH}
     \end{subfigure}
     \hfill
     \begin{subfigure}[b]{0.24\textwidth}
         \centering
         \includegraphics[width=\textwidth]{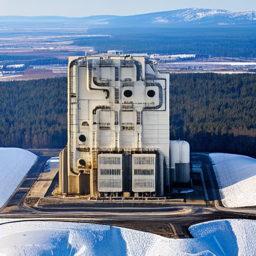}
         \caption{Winter}
         \label{fig:SWGSH}
     \end{subfigure}
     \caption{Visualization of different seasonality through prompt engineering with default 10 (top) and high 15 (bottom) guidance scale. In addition to the visual input modality, the text prompt ``Satellite image of a nuclear power plant in [season]'' was used.}
        \label{fig:Season}
\end{figure}

While most features like plant structure, time-of-day, cloud coverage, and level of activity onsite can be adjusted within the game engine, we decided to capture seasonality through text prompts and different guidance scales of the input modalities, rather than implementing a comparatively costly additional step in the game engine that would probably only lead to mediocre results. The output is shown in \autoref{fig:Season}.
While most of the results are convincing, in the case of using a high guidance scale of the text prompt for ``fall'' and ``winter'', the model generates a background from a low-elevation viewpoint even though the text prompt was ``\textbf{Satellite} image of a nuclear power plant in fall/winter''.
This shows that the model is biased towards typical viewpoints, as satellite imagery is underrepresented in the training data. So the guidance scale of the text prompt shouldn't be set too high in order to prevent the model from generating incoherent results.

\begin{figure}

     \begin{subfigure}[b]{0.24\textwidth}
         \centering
         \includegraphics[width=\textwidth]{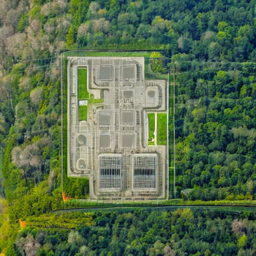}
     \end{subfigure}
     \hfill
          \begin{subfigure}[b]{0.24\textwidth}
         \centering
         \includegraphics[width=\textwidth]{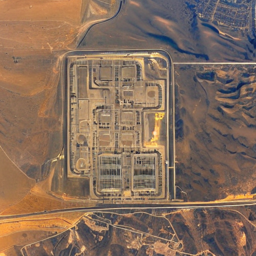}
     \end{subfigure}
     \hfill
          \begin{subfigure}[b]{0.24\textwidth}
         \centering
         \includegraphics[width=\textwidth]{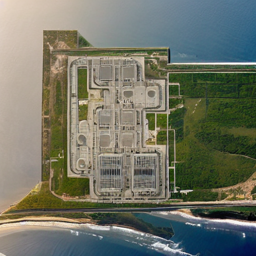}
     \end{subfigure}
     \hfill
     \begin{subfigure}[b]{0.24\textwidth}
         \centering
         \includegraphics[width=\textwidth]{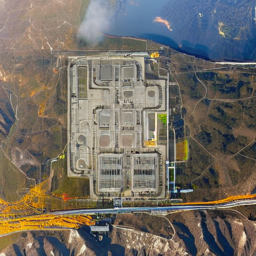}
     \end{subfigure}
     
     \begin{subfigure}[b]{0.24\textwidth}
         \centering
         \includegraphics[width=\textwidth]{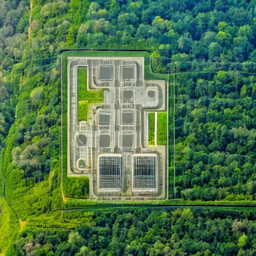}
         \caption{"in a forest"}
         \label{fig:EF}
     \end{subfigure}
     \hfill
          \begin{subfigure}[b]{0.24\textwidth}
         \centering
         \includegraphics[width=\textwidth]{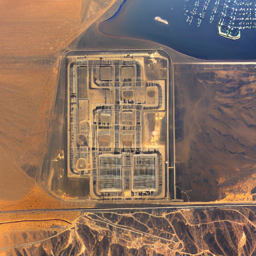}
         \caption{"in the desert"}
         \label{fig:ED}
     \end{subfigure}
     \hfill
          \begin{subfigure}[b]{0.24\textwidth}
         \centering
         \includegraphics[width=\textwidth]{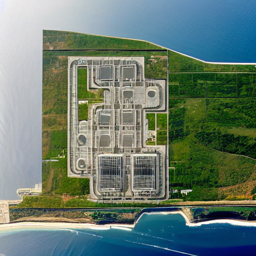}
         \caption{"by a coastline"}
         \label{fig:EC}
     \end{subfigure}
     \hfill
     \begin{subfigure}[b]{0.24\textwidth}
         \centering
         \includegraphics[width=\textwidth]{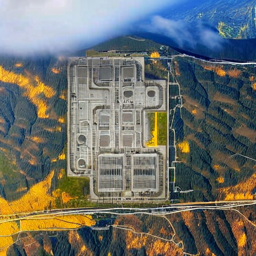}
         \caption{"in the mountains"}
         \label{fig:EM}
     \end{subfigure}
     \caption{Visualization of different environments through prompt engineering with default 10 (top) and high 15 (bottom) guidance scale. In addition to the visual input modality, the text prompt ``Satellite image of a nuclear power plant in/by/on [environment]'' was used.}
        \label{fig:Env}
\end{figure}

The same way seasonality can be changed, the environment can be specified by the text prompt, as shown in \autoref{fig:Env}.

The problem of disappearing details, such as the steam from the cooling towers, clouds, or cars in the parking lot, can easily be solved by extracting those details from the game engine render and adding them to the image synthesized by the CoAdapter model, as illustrated in \autoref{fig:Overview}. This process can be automated, as each object can be separately rendered in the game engine. More general image features such as the lighting mood at different times-of-day, as illustrated in \autoref{fig:Time}, can be added to the synthesized image by creating a blend between both images.

\begin{figure}     
     \begin{subfigure}[b]{0.32\textwidth}
         \centering
         \includegraphics[width=\textwidth]{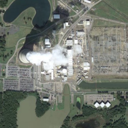}
         \caption{Satellite image of a real nuclear power plant}
         \label{fig:RefReal}
     \end{subfigure}
     \hfill
          \begin{subfigure}[b]{0.32\textwidth}
         \centering
         \includegraphics[width=\textwidth]{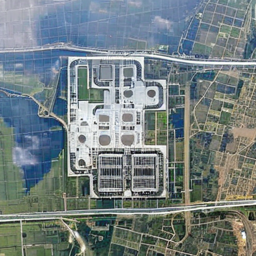}
         \caption{Synthesized result from color map of reference image}
         \label{fig:MLEKD}
     \end{subfigure}
     \hfill
     \begin{subfigure}[b]{0.32\textwidth}
         \centering
         \includegraphics[width=\textwidth]{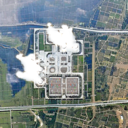}
         \caption{Final image with details from game engine render}
         \label{fig:RefColor}
     \end{subfigure}
        \caption{Synthesized results of canny edge with visual input modality of real satellite image and reinserted details}
        \label{fig:RefImage}
\end{figure}

By using the stylistic features of a real satellite image along with the text prompt and content-related features of a synthetic rendering, it is possible to generate a realistic image that includes the desired elements presented in a more realistic style, as illustrated in \autoref{fig:RefImage}.

\section{Discussion}

\subsection{Method Limitations and Desiderata}
Despite the advantages of combining game engines and machine learning, there are still limitations to our method. While we showed that arbitrary views of procedurally generated layouts of pre-modelled structures can result in a large number of synthetic satellite images, the quality of which can be greatly improved through the multimodally-guided CoAdapter model, and illustrated how details can be reinserted into the final images, on closer inspection the images are still not entirely realistic. This is due to the fact that the reinserted details are limited by progress on game engine graphics, which would need to be manually optimized for each image generated by the CoAdapter model, for example by adjusting the details to the specific lighting.
%
And although the CoAdapter model can narrow the domain gap between game engine renderings and real images, it lacks the ability to produce photo-realistic results. Streets are often simplified, structures asymmetrical, and shadows sometimes inconsistent -- all indications that the image is synthetic.


In summary, what has been achieved so far is a procedure for generating synthetic satellite images of a notional nuclear power plant. Starting with a 3D game engine model of the plant composed of its basic elements, the rendering can be modified to account for variable conditions pertaining to both the environment being imaged (time-of-day, cloud coverage, level of activity onsite such as steam plumes and number of cars) as well as satellite image acquisition (camera distance, off-nadir angle). The images are then further enhanced in terms of appearance and content via the machine learning-based CoAdapter model. Thanks to its multiple input modalities, the text prompt given to the CoAdapter model can further be used to modify the images to account for variable seasonality and surrounding environment. The realism of the resulting imagery can also be enhanced by providing the CoAdapter model with a real satellite image. Since the 3D model layout and the image acquisition and environmental conditions are configurable, a large set of images exhibiting high variability can be generated with minimal manual labor.
Determining the ``best'' combination of input modalities and guidance scale and manually re-inserting details take several minutes per image -- steps that constitute the largest overhead. For future work, the detail re-insertion process can be automated by rendering the details only and adding them automatically to all variations of input modalities so that the ``best'' result of the $2^4=16$ possible input modality combinations (canny, depth, sketch, and color of reference photo) can be chosen from multiple proposals that can be generated in under three minutes. 



For the purposes of developing materials for a tabletop exercise a number of additional improvements are needed to capture the full range of variability in image acquisition and environmental conditions. First, we need to be able to simulate imaging spatial resolution to reflect inherent finiteness, but also to capture degradation effects with the variation of camera distance, off-nadir angle, and environmental conditions such as time-of-day and weather and atmospheric phenomena.
%
%
Second, we want to be able to generate imagery in different spectral bands, including the infrared band for the detection of thermal indicators of nuclear activity and the microwave band, which enables visibility through clouds and during night-time. Synthetic aperture radar imagery is also potent and would be interesting to synthesize \cite{Moric2022}.
Finally, for the assessment of the power of satellite imagery in enabling change detection and activity tracking, we need to be able to generate an image sequence over a given temporal window at specific intervals conforming to actual Earth-observation satellite operation, i.e., capturing satellite orbital trajectory over the temporal window as well as the image acquisition frequency and the corresponding variation in environmental conditions (which can in turn influence image quality) over the acquisition times. The scope of the required imagery to assess applicability to change detection and activity tracking is expected to increase in this case to encompass \begin{enumerate*}[label=\textit{(\alph*)}]\item a wider swath of environing road networks as well as other types of infrastructure relevant to nuclear material production, e.g., mines and ports; \item longer time intervals corresponding to the duration of the nuclear activities of interest; \item different bands and spatial and temporal resolutions as required by the nuclear activity indicators of interest.\end{enumerate*}



\subsection{Ethical and Political Considerations}
    \paragraph{Large-scale data collection and analysis:} The explosion in data, compute power, and algorithms in recent years has prompted interest in agreement monitoring in the nuclear context relying on open sources and machine learning-based analysis. Work by \citeauthor{Gastelum2018} \cite{Gastelum2018} and \citeauthor{Feldman2018} \cite{Feldman2018} and other work featured in the same Journal of Nuclear Materials Management 2018 issue exemplify the trend in safeguards research, where the state-level concept is often the pretext justifying large-scale data collection and analysis. However, there is dire need to resolve trust issues between states based on a solid common understanding, to say nothing of the trust issues between human and machine. The tabletop exercise attempts to flesh out some of the underlying issues of concern.
    
    \paragraph{Bias perpetuation:} In projects involving the generation of synthetic imagery of a given object of interest, pre-existing notions of the object of interest tend to influence the procedure and in turn, the generated dataset. So far, the procedure is guided by prenotions about the appearance of a nuclear power plant in three obvious ways. First, the construction of the 3D model of a notional plant is inspired by domain knowledge of the appearance of real plants. Second, a real image of an existing nuclear power plant is used as reference for the CoAdapter model to enhance the realism of the generated synthetic imagery. Third, when the CoAdapter is guided by a text prompt that refers to a `nuclear power plant', the underlying machine learning procedure powering the text-to-image generation produces an output that is ultimately influenced by training data in the form of existing imagery representing the concept `nuclear power plant'. Recalling to mind the lack of sufficient, representative real imagery of nuclear power plants, this training dataset can safely be assumed to be small, limiting the variability of the generated synthetic imagery. Together, these three factors result in the output imagery manifesting a degree of bias towards pre-existing notions that ideally one would like to reduce, if not eliminate, in the context of a tabletop exercise that seeks to assess the potential and limitations of the use of satellite imagery in a controlled setup that guards against the participants' being influenced by their prenotions.

\section{Conclusion}


Our demonstration showcases how the discrepancy between 3D game engine renders and real satellite imagery can be reduced by means of multimodally-guided image synthesis. Additionally, our approach enables the creation of sizable high-variability datasets in a brief span of time, making it particularly useful in cases where comprehensive satellite imagery is rare, which is the case for nuclear power plants. With this work, we have laid the groundwork for the assessment of the long-term potential of satellite imagery for nuclear activity monitoring and verification purposes.


With advances in machine learning and with the growth in high-quality real imagery datasets for training, it's to be expected that synthetic Earth-observation imagery generation will improve in quality and that synthetic imagery will proliferate. And while synthetic imagery has countless good uses, it's to be expected that malicious actors would use doctored images to stage hoaxes or discredit reality, in both cases to mislead adversaries, entities tasked with monitoring and verification, or publics. And while `deepfake geography' is a milennia-old practice, novel artificial intelligence (AI)-generated deepfake satellite imagery introduces new challenges \cite{Zhao2021}. For example, satellite imagery's characteristically low resolution compared to other forms of imagery such as photography works to the favor of deepfakes, since the latter are then easier to produce and deemed authentic by the observer. The convincing power of deepfake satellite imagery can also be attributed to the level of complexity and expense observers rightfully ascribe to the long, intricate, and costly process of satellite imagery generation -- a process embedded in a sociotechnical complex mediated by science, economics, politics, \ldots \cite{Verge-DeepfakeSatImagery}. As a form of disinformation, undetected deepfake satellite imagery could have serious consequences for international peace and security, and current governance structures are concernedly lagging in keeping pace with AI developments.
    

\section*{Acknowledgments}

The authors thank the German Foundation for Peace Research (DSF) for their support of this research effort as part of our ongoing project ``Citizen-based Monitoring for Peace \& Security in the Era of Synthetic Media and Deepfakes.''


\printbibliography




  \end{document}